\documentclass[11pt]{article}

\usepackage{amsmath}
\usepackage{amssymb}
\usepackage{amsfonts}
\usepackage{amsthm}

\usepackage{graphicx}
\usepackage{tikz}
\usepackage{hyperref}
\usepackage{geometry}

\usepackage{bm}
\usepackage{mathtools}
\usepackage{enumitem}
\usepackage{cleveref}

\usepackage{float}

\usepackage[numbers,sort&compress]{natbib}

\newtheorem{proposition}{Proposition}

\usetikzlibrary{arrows.meta,calc,positioning}

\title{Gauge Freedom and Metric Dependence in Neural Representation Spaces}

\author{
Jericho Cain\\
Physics Department, Portland Community College, Portland, OR, USA\\
\texttt{jericho.cain@gmail.com}\\
ORCID: 0000-0003-4731-9142
}

\begin{document}
\hbadness=10000

\maketitle

\begin{abstract}

Neural network representations are often analyzed as vectors in a fixed
Euclidean space. However, their coordinates are not uniquely defined.
If a hidden representation is transformed by an invertible linear map, the
network function can be preserved by applying the inverse transformation to
downstream weights. Representations are therefore defined only up to
invertible linear transformations.
We study neural representation spaces from this geometric viewpoint and
treat them as vector spaces with a gauge freedom under the general linear
group. Within this framework, commonly used similarity measures such as cosine
similarity become metric-dependent quantities whose values can change under
coordinate transformations that leave the model function unchanged.
This provides a common interpretation for several observations in the
literature, including cosine-similarity instability, anisotropy in
embedding spaces, and the appeal of representation comparison methods such
as SVCCA and CKA. Experiments on multilayer perceptrons and convolutional
networks confirm that inserting invertible transformations into trained models can
substantially distort cosine similarity and nearest-neighbor structure
while leaving predictions unchanged.
These results indicate that analysis of neural representations should focus
either on quantities that are invariant under this gauge freedom or on
explicitly chosen canonical coordinates.

\end{abstract}
\section{Introduction}

Modern neural networks operate by transforming inputs into sequences of
high-dimensional vector representations.
These representations appear throughout machine learning systems, including
word embeddings, hidden states in transformer models, and latent variables
in autoencoders.
Because of this ubiquity, a large body of work attempts to understand neural
models by analyzing the geometry of these vector representations \cite{Bengio2012}.

A common assumption in such analyses is that the coordinates of the
representation vectors carry intrinsic geometric meaning.
Similarity between representations is frequently measured using cosine
similarity or Euclidean distance, and structural properties of the
representation space are studied through principal components, clustering,
or subspace comparisons.

The coordinates used to describe neural representations are not intrinsic.
If $h(x)\in\mathbb{R}^d$ is a hidden representation and $W$ is a linear
readout, then for any $D\in\mathrm{GL}(d)$ (the group of invertible
$d\times d$ matrices) the transformation
\[
h'(x)=Dh(x),
\qquad
W'=WD^{-1}
\]
leaves the network function unchanged since $W'h'(x)=Wh(x)$.
Representations are defined only up to the action of
$\mathrm{GL}(d)$. We refer to this freedom as a gauge symmetry of representation space.
Geometric quantities tied to a coordinate choice, including cosine
similarity, need not remain invariant under this transformation.

For example, cosine similarity is often used to measure semantic similarity
between embeddings, yet cosine similarity depends on the metric structure
of the space.
Under a linear transformation $h \mapsto Dh$, cosine similarity between
representations generally changes even though the encoded information
remains the same.

Recent work has reported related phenomena in several contexts.
Steck et al.~\cite{Steck2024} show that cosine similarity applied to learned
embeddings in matrix factorization models can yield arbitrary results due
to scaling freedoms in the factorization.
Ethayarajh~\cite{Ethayarajh2019} studies the anisotropy of contextualized
embedding spaces and shows that cosine similarity is strongly influenced
by dominant directions in the representation distribution.
Ai~\cite{Ai2026} argues that cosine similarity captures only angular
relationships between vectors and may fail to reflect other meaningful
relations in embedding spaces.
Meanwhile, representation comparison methods such as SVCCA and CKA have been
proposed to compare neural representations while remaining insensitive to
certain linear transformations~\cite{Raghu2017,Kornblith2019}.

We analyze representation spaces as vector spaces defined only up to
invertible linear transformations and track how this affects metric-based
geometry.
The experiments validate the central claim: cosine similarity and
nearest-neighbor structure can shift substantially under gauge transforms
that preserve the model function.

\section{Representation Spaces}

For a network with parameters $\theta\in\mathbb{R}^p$, let
\[
f_\theta:\mathcal{X}\to\mathcal{Y}
\]
denote the overall input-output map.
At an intermediate layer $\ell$, the network produces a hidden
representation
\[
h_\ell(x;\theta)\in V,
\]
where $V=\mathbb{R}^d$ is a $d$-dimensional vector space.
We refer to $V$ as the representation space of layer $\ell$.

For a dataset $\{x_i\}_{i=1}^n$, the representations produced by the
network form a collection of vectors

\[
h_\ell(x_1;\theta),\dots,h_\ell(x_n;\theta) \in V.
\]

These vectors can be arranged into a representation matrix

\[
H =
\begin{pmatrix}
h_\ell(x_1;\theta) & \cdots & h_\ell(x_n;\theta)
\end{pmatrix}
\in \mathbb{R}^{d \times n}.
\]

Many empirical analyses of neural networks examine the geometry of these
representations.
Examples include similarity comparisons between embeddings, clustering
structure in hidden states, and dimensionality reduction techniques such
as principal component analysis.

Throughout this work we treat the representation space $V$ as a vector
space equipped with an inner product

\[
\langle u,v\rangle = u^\top v.
\]

This inner product induces the Euclidean metric

\[
\|u\| = \sqrt{u^\top u}.
\]

Common similarity measures such as cosine similarity are defined with
respect to this metric.
However, as we show in the next section, the coordinates of representation
vectors are not uniquely defined.

\section{Gauge Freedom of Representations}

The coordinates of neural representations are not unique.
Consider a hidden representation

\[
h(x) \in V = \mathbb{R}^d.
\]

Suppose the next layer of the network computes

\[
y = W h(x),
\]

where \(W\) is a linear map acting on the representation.

Let \(D \in \mathrm{GL}(d)\) be any invertible matrix and define a transformed
representation

\[
\tilde h(x) = D h(x).
\]

If the downstream weights are adjusted to

\[
\tilde W = W D^{-1},
\]

then the network output can be written as

\[
y = \tilde W \tilde h(x).
\]

Thus the same network function can be expressed using the transformed
representation \(\tilde h(x)\) provided the downstream linear map is
adjusted accordingly.

\begin{proposition}[Representation gauge symmetry]
\label{prop:gauge}
Let \(h(x)\in\mathbb{R}^d\) be a hidden representation and suppose a
subsequent linear layer computes \(y = W h(x)\).
For any invertible matrix \(D\in\mathrm{GL}(d)\), define

\[
\tilde h(x) = D h(x),
\qquad
\tilde W = W D^{-1}.
\]

Then

\[
\tilde W \tilde h(x) = W h(x),
\]

so the network function remains unchanged.
Thus the coordinates of representation space are defined only up to the
action of \(\mathrm{GL}(d)\).
\end{proposition}

Consequently, two representation systems related by an invertible linear
transformation

\[
h(x) \mapsto D h(x)
\]

encode identical information for the purposes of the network computation.

The vectors \(h(x)\) are therefore defined only up to a change of basis in
\(V\).
We refer to this symmetry as a gauge freedom of representation space.
Formally, representation vectors are defined modulo the action of
\(\mathrm{GL}(d)\).

While this freedom does not affect the function computed by the network,
it can alter geometric properties of the representation coordinates.
Quantities that depend on the metric structure of the space, such as cosine
similarity, may change under such transformations.
Analysis of representation geometry should therefore distinguish
gauge-dependent quantities from gauge-invariant ones.

\section{Metric Structure and Similarity Measures}

The gauge freedom described in the previous section implies that the
coordinates of representation vectors are not uniquely defined.
Consequently, geometric quantities that depend on the metric structure
of the representation space may vary under invertible linear
transformations.

Let $u,v \in V$ be two representation vectors.
Cosine similarity is commonly used to measure their similarity:

\[
\cos(u,v) =
\frac{u^\top v}{\|u\|\|v\|}.
\]

This quantity implicitly assumes the Euclidean metric on $V$.

Now consider a gauge transformation

\[
\tilde u = D u,
\qquad
\tilde v = D v,
\]

where $D \in \mathrm{GL}(d)$.
The inner product between the transformed vectors becomes

\[
\langle \tilde u,\tilde v\rangle
=
u^\top D^\top D v .
\]

The transformation induces a new inner product

\begin{equation}
\langle u,v\rangle_D
=
u^\top G v,
\qquad
G = D^\top D .
\label{eq:metric_pullback}
\end{equation}

Geometrically, the matrix $G$ acts as a metric tensor on the
representation space.
Under this metric, cosine similarity becomes

\[
\cos_D(u,v)
=
\frac{u^\top G v}
{\sqrt{u^\top G u}\sqrt{v^\top G v}}.
\]

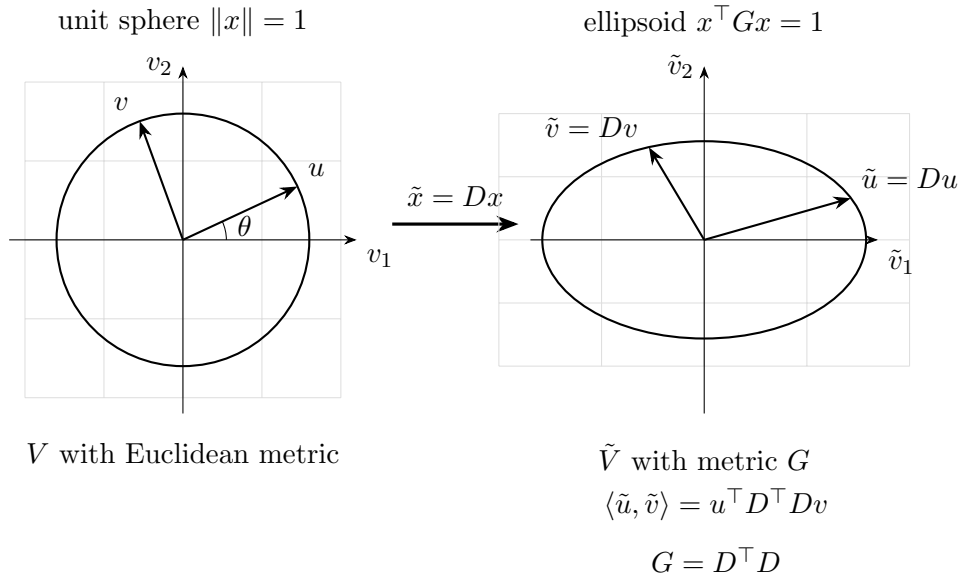
\begin{figure}[ht]
\centering
\begin{tikzpicture}[scale=1.05, >=Stealth]

  \begin{scope}

    \draw[gray!25, thin] (-2,-2) grid (2,2);

    \draw[->] (-2.2,0) -- (2.2,0) node[below right] {$v_1$};
    \draw[->] (0,-2.2) -- (0,2.2) node[left] {$v_2$};

    \draw[thick] (0,0) circle (1.6);

    \coordinate (O) at (0,0);
    \coordinate (u) at ({1.6*cos(25)},{1.6*sin(25)});
    \coordinate (v) at ({1.6*cos(110)},{1.6*sin(110)});

    \draw[thick,->] (O) -- (u) node[above right] {$u$};
    \draw[thick,->] (O) -- (v) node[above left] {$v$};

    \draw (0.55,0) arc (0:25:0.55);
    \node at (0.78,0.18) {$\theta$};

    \node[below] at (0,-2.45) {$V$ with Euclidean metric};
    \node[above] at (0,2.45) {unit sphere $\|x\|=1$};

  \end{scope}

  \draw[very thick,->] (2.65,0.2) -- (4.25,0.2);
  \node at (3.45,0.55) {$\tilde{x}=Dx$};

  \begin{scope}[xshift=6.6cm]

    \draw[gray!25, thin, xscale=1.3, yscale=0.8] (-2,-2) grid (2,2);

    \draw[->] (-2.2,0) -- (2.2,0) node[below right] {$\tilde v_1$};
    \draw[->] (0,-2.2) -- (0,2.2) node[left] {$\tilde v_2$};

    \draw[thick] (0,0) ellipse (2.05 and 1.25);

    \coordinate (Ou) at (0,0);
    \coordinate (tu) at ({1.28*1.6*cos(25)},{0.78*1.6*sin(25)});
    \coordinate (tv) at ({1.28*1.6*cos(110)},{0.78*1.6*sin(110)});

    \draw[thick,->] (Ou) -- (tu) node[above right] {$\tilde u=Du$};
    \draw[thick,->] (Ou) -- (tv) node[above left] {$\tilde v=Dv$};

    \node[above] at (0,2.45) {ellipsoid $x^\top G x = 1$};
    \node[below] at (0,-2.45) {$\tilde V$ with metric $G$};

    \node[align=left] at (0.15,-3.35) {$\langle \tilde u,\tilde v\rangle
      = u^\top D^\top D v$};
    \node[align=left] at (0.15,-4.05) {$G=D^\top D$};

  \end{scope}

\end{tikzpicture}
\caption{A linear distortion $\tilde x = D x$ maps the Euclidean unit sphere to an ellipsoid. The induced inner product is $G=D^\top D$, so cosine similarity computed after the transformation corresponds to angular similarity in the distorted metric.}
\label{fig:metric-distortion-ellipse}
\end{figure}

Cosine similarity is not invariant under general gauge transformations of the
representation space.
Linear distortion changes angular relationships by inducing a new metric
tensor (Figure~\ref{fig:metric-distortion-ellipse}).
Two embedding systems related by an invertible linear transformation may
represent the same information while producing different cosine
similarity values.

\subsection{Whitening as a Canonical Gauge}

A particularly important case arises when the transformation removes
anisotropy in the representation distribution.

Let $\Sigma$ denote the covariance matrix of hidden states,

\[
\Sigma = \mathbb{E}[h h^\top].
\]

Defining the whitening transformation

\[
D = \Sigma^{-1/2}
\]

maps the covariance to the identity matrix.
In the transformed coordinates

\[
\tilde h = D h
\]

the representation distribution becomes isotropic,

\[
\mathbb{E}[\tilde h \tilde h^\top] = I.
\]

In these coordinates, cosine similarity corresponds to angular similarity
in a space where the representation distribution has unit covariance.
Equivalently, cosine similarity in the original coordinates can be
interpreted as angular similarity under the Mahalanobis metric induced by
$\Sigma$.

Within this setup, whitening is a gauge choice that fixes the metric
structure of the representation space.

\subsection{Implications for Embedding Similarity}

Recent empirical work has observed that cosine similarity can behave
unexpectedly in embedding spaces.

Steck et al.~\cite{Steck2024} analyze cosine similarity in matrix
factorization models and show that the learned embeddings may contain
scaling freedoms that allow arbitrary rescaling of latent dimensions.
These rescalings leave the model predictions unchanged while producing
different cosine similarity values between embeddings.

Similarly, Ai~\cite{Ai2026} argues that cosine similarity captures only
angular alignment between vectors and may fail to reflect other
relationships between embeddings.

Within the geometric framework developed here, these observations have a
direct explanation.
Invertible linear transformations alter the metric structure of the
representation space while preserving the underlying information encoded
by the representations.
Similarity measures that depend on the metric, such as cosine similarity,
become coordinate dependent.
Representation analysis should prioritize quantities that remain invariant
under gauge freedom.

\section{Feature Directions and Subspaces}

Many analyses of neural representations attempt to identify interpretable
features encoded within hidden states.  These features are often modeled as
directions in the representation space.

Let $h(x) \in V = \mathbb{R}^d$ denote the representation of an input $x$.
Suppose that the representation can be approximated as a linear combination
of feature vectors,

\[
h(x) \approx F a(x),
\]

where

\[
F =
\begin{pmatrix}
f_1 & f_2 & \cdots & f_k
\end{pmatrix}
\in \mathbb{R}^{d \times k}
\]

is a matrix whose columns correspond to feature directions and
$a(x) \in \mathbb{R}^k$ are feature activations.

In this view, each feature corresponds to a direction in representation
space, and the representation vector encodes the strengths with which
these features are activated.

\subsection{Feature Packing and Superposition}

In many neural networks the number of potential features exceeds the
dimension of the representation space.
Consequently, multiple features may be encoded within overlapping
subspaces of the representation space.

This phenomenon is often described as feature superposition
\cite{Elhage2022}.
In the linear model above, superposition occurs when the feature matrix
$F$ has more columns than the ambient dimension of the representation
space.

The geometry of feature interactions can be characterized by the
Gram matrix of the feature directions,

\[
G_F = F^\top F .
\]

When the columns of $F$ are orthogonal, the Gram matrix is diagonal and
features can be read out independently.
When the columns are not orthogonal, feature activations interact through
the off-diagonal entries of $G_F$, producing interference between
features.

If a representation is written as
\[
h = F a
\]
with a sparse activation vector $a$, then recovering the active features
depends on the Gram matrix $G_F$ being close to diagonal on the indices
where $a_i \neq 0$.
The geometry of superposition is controlled not only by the number of
features, but by the metric-dependent overlap structure among the feature
directions.

Feature superposition is a geometric property of representation space rather
than a property of individual neurons.

\subsection{Gauge Transformations of Feature Bases}

Under a gauge transformation of the representation space,

\[
h \mapsto D h,
\qquad
D \in \mathrm{GL}(d),
\]

the feature matrix transforms as

\[
F \mapsto D F .
\]

The Gram matrix of the features becomes

\[
F^\top F
\mapsto
F^\top D^\top D F .
\]

The pairwise angles between feature directions depend on the metric
structure induced by the gauge choice.
Two representation systems related by an invertible linear transformation
may still contain identical information while exhibiting different
apparent feature geometries.

Many geometric properties of representations depend on the chosen
coordinates for representation space.

\subsection{Representation Similarity Across Networks}

A related problem arises when comparing representations produced by
different neural networks.
Given two representation matrices

\[
H_1 , H_2 \in \mathbb{R}^{d \times n},
\]

direct comparison of their coordinates is generally not meaningful
because the two representation spaces may differ by an invertible linear
transformation.

Several methods have been proposed to address this issue.
Canonical correlation analysis (CCA) and its variants align
representations by searching for linear projections that maximize
correlation between corresponding activations.
More recently, centered kernel alignment (CKA) compares representations
using Gram matrices of example similarities.

Within the geometric framework developed here, these methods can be viewed
as attempts to reduce the dependence of representation comparisons on the
coordinate realization of representation space.  Rather than comparing
representation vectors directly, approaches such as CCA and CKA compare
subspaces or similarity structures derived from the representations.
These quantities tend to remain more stable under invertible linear
transformations of the representation basis.  From this perspective,
representation comparison methods can be interpreted as approximations to
gauge-invariant observables of the representation space.

\section{Representation Dynamics and Local Geometry}

The previous sections focused on the geometric structure of representation
spaces at a fixed set of network parameters.
During training, however, representations evolve as the parameters of the
network are updated.
Understanding how parameter updates affect representations provides a
useful perspective on the dynamics of learning.

Let $\theta \in \mathbb{R}^p$ denote the parameters of a neural network and
consider the representation at layer $\ell$

\[
h_\ell(x;\theta) \in \mathbb{R}^d .
\]

For small parameter changes $\delta \theta$, the corresponding change in the
representation can be approximated using a first-order Taylor expansion,

\begin{equation}
\delta h_\ell(x)
\approx
J_\ell(x) \, \delta \theta ,
\label{eq:linear_rep_dynamics}
\end{equation}

where

\[
J_\ell(x)
=
\frac{\partial h_\ell(x;\theta)}{\partial \theta}
\in \mathbb{R}^{d \times p}
\]

is the Jacobian of the representation with respect to the model parameters.

\subsection{Induced Metric on Parameter Space}

Equation~(\ref{eq:linear_rep_dynamics}) relates parameter updates to
representation changes.
Using the Euclidean norm in representation space, the squared magnitude of
the representation change becomes

\begin{equation}
\|\delta h_\ell(x)\|^2
\approx
\delta\theta^\top (J_\ell^\top J_\ell)\, \delta\theta .
\label{eq:rep_change_metric}
\end{equation}

The matrix

\[
G_\ell = J_\ell^\top J_\ell
\]

defines a positive semidefinite quadratic form on parameter space.
Geometrically, this matrix acts as a metric tensor describing how changes
in parameters affect the representation at layer $\ell$.

This construction is closely related to the idea of pullback metrics in
differential geometry.
The Jacobian maps directions in parameter space to directions in
representation space, and the metric $G_\ell$ measures parameter changes in
terms of the resulting change in representations.

\subsection{Interpretation for Training Dynamics}

The metric $G_\ell$ provides a local description of how sensitive the
representation is to parameter updates.
Directions in parameter space corresponding to large eigenvalues of
$G_\ell$ produce large changes in the representation, while directions
corresponding to small eigenvalues produce little change.

Representation geometry is tied to optimization through this metric.
Gradient-based learning updates parameters in directions determined by the
loss function, but the resulting change in the representation depends on
how these parameter directions are mapped through the Jacobian of the
network.

Consequently, different layers may experience different effective
geometries during training depending on the structure of their Jacobians.

\subsection{Gauge Dependence}

As in earlier sections, representation geometry depends on the coordinate
system used to describe the representation space.
Under a gauge transformation
\[
h_\ell \mapsto D h_\ell ,
\]
the Jacobian transforms as
\[
J_\ell \mapsto D J_\ell .
\]

The pullback metric on parameter space becomes
\[
G_\ell \mapsto J_\ell^\top D^\top D J_\ell .
\]

A complementary perspective considers the covariance of representation
changes.
If parameter updates have covariance $\Omega$, then under the linearization
\eqref{eq:linear_rep_dynamics} the covariance of representation changes is
approximately
\[
\mathrm{Cov}(\delta h_\ell)
\approx
J_\ell \Omega J_\ell^\top .
\]

Training induces a layer-dependent anisotropy in representation
space: directions associated with large eigenvalues of
$J_\ell \Omega J_\ell^\top$ are directions along which optimization can
move representations most easily.

This connects the geometry of training dynamics to the metric ambiguity
discussed earlier.
A change of gauge alters the metric used to describe representation space,
while optimization itself induces preferred directions of motion inside
that space.

\section{Experiments}

We conduct a set of controlled experiments to illustrate the geometric
properties of representation space discussed in the preceding sections.
The goal of these experiments is not to improve model performance but to
demonstrate that commonly used similarity measures depend on the choice of
representation coordinates.

In each experiment we apply an invertible linear transformation
\[
h' = D h
\]
to a hidden representation $h$ and compensate the final linear classifier
by replacing $W$ with
\[
W' = W D^{-1}.
\]
This transformation preserves the network function since
\[
W' h' = W h.
\]
Predictions remain unchanged while the representation coordinates are
altered. This procedure isolates the geometric effect of a gauge
transformation of representation space: the model function is held fixed
while the coordinate realization changes. These transformations
correspond exactly to the gauge symmetry described in
Proposition~\ref{prop:gauge}.
\subsection{Digits Dataset: Sanity Check}

We first trained a two-layer multilayer perceptron on the
\texttt{scikit-learn} Digits dataset and extracted hidden layer
representations from the test split.

Applying the gauge transformation described above produced identical model
predictions up to numerical precision.
The maximum difference in logits after the transformation was
approximately $1.5\times10^{-5}$ and prediction agreement was $1.0$.

Despite this functional invariance, the geometry of the representation
space changes substantially.
The mean absolute change in pairwise cosine similarity between hidden
states is

\[
\text{mean }|\Delta \cos| = 0.1328 .
\]

\begin{figure}[ht]
\centering
\includegraphics[width=\linewidth]{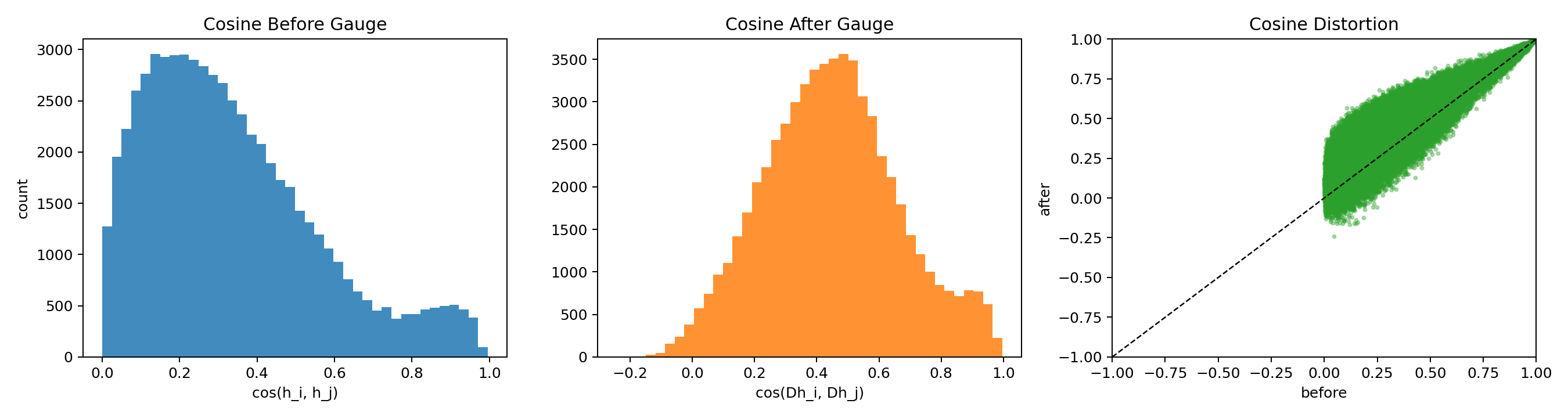}
\caption{Pairwise cosine similarities before and after a gauge transformation
in the Digits experiment. The model predictions remain identical, but cosine
similarities between representations change substantially.}
\label{fig:digits_cosine_distortion}
\end{figure}

Figure~\ref{fig:digits_cosine_distortion} illustrates this effect.
Although the network function remains unchanged, the distribution of cosine
similarities shifts noticeably and individual pairwise similarities can
change significantly.

Nearest-neighbor structure also changed.
For cosine-based retrieval using $k=10$ neighbors, the mean Jaccard overlap
between neighbor sets before and after the transformation was

\[
\text{Jaccard@10} = 0.7209 .
\]

Roughly $28\%$ of nearest neighbors changed even though the model
function was identical.

\subsection{CIFAR-10 Convolutional Network}

To verify that the phenomenon is not specific to small toy models, we
repeated the experiment using a small convolutional neural network trained
on the CIFAR-10 dataset.

The network reached a test accuracy of $0.6823$ after five training
epochs.
We extracted penultimate-layer representations and applied the same gauge
transformation with the compensating inverse in the classifier.

As in the Digits experiment, the model function remained invariant.
Prediction agreement was $1.0$ and the maximum logit difference was on the
order of $10^{-5}$.

The average cosine distortion was smaller than in the Digits experiment,

\[
\text{mean }|\Delta \cos| = 0.0501 ,
\]

but nearest-neighbor structure still changed significantly.

\begin{figure}[ht]
\centering
\includegraphics[width=\linewidth]{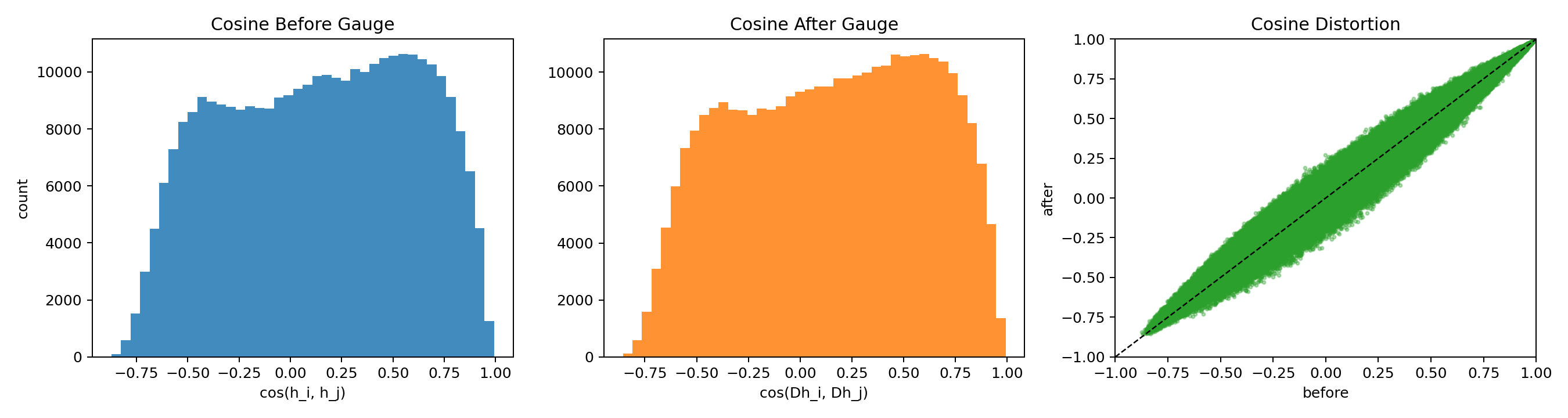}
\caption{Cosine distortion for CIFAR-10 representations under a gauge
transformation. Predictions remain unchanged, but pairwise cosine similarity
between hidden states is altered.}
\label{fig:cifar_cosine_distortion}
\end{figure}

Cosine geometry still shifts under gauge transforms
(Figure~\ref{fig:cifar_cosine_distortion}).
Although the magnitude of the cosine changes is smaller than in the
Digits experiment, many pairwise similarities are systematically shifted
under the gauge transformation.

The mean neighbor overlap was

\[
\text{Jaccard@10} = 0.7228 .
\]

Even moderate distortions in cosine similarity can produce noticeable
changes in nearest-neighbor relationships.

\subsection{Gauge Strength Sweep}

We constructed a family of gauge transformations with increasing condition
number
\[
\kappa = \frac{\sigma_{\max}(D)}{\sigma_{\min}(D)} .
\]

For each value of $\kappa$ we applied the transformation to the CIFAR-10
representations and report mean $|\Delta \cos|$, nearest-neighbor stability
(Jaccard@10), and the top-1 neighbor flip rate.

\begin{figure}[ht]
\centering
\includegraphics[width=\linewidth]{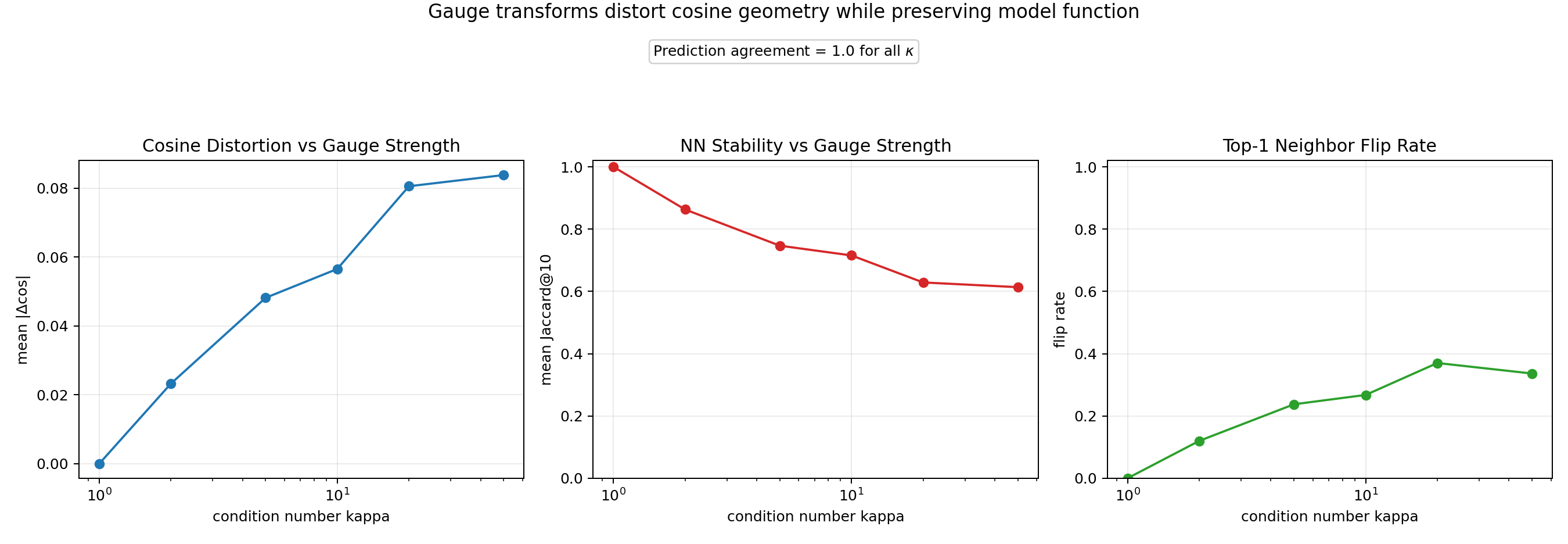}
\caption{Effect of gauge strength on cosine similarity and nearest-neighbor
structure in CIFAR-10 representations.  Increasing the condition number
$\kappa$ of the transformation produces larger cosine distortions and
reduces neighbor stability, even though the network function remains
unchanged.}
\label{fig:cifar_kappa_sweep}
\end{figure}

The results are shown in Figure~\ref{fig:cifar_kappa_sweep}, which isolates
how increasing gauge strength affects representation geometry while the
model function remains unchanged. For all values of $\kappa$, prediction
agreement remained $1.0$ and the network function was unchanged up to
numerical precision.

At $\kappa = 1$ the transformation is orthogonal and the cosine similarity
structure is preserved.
As $\kappa$ increases, cosine distortion grows and nearest-neighbor
structure becomes less stable.

For example, at $\kappa = 20$ we observe

\[
\text{mean }|\Delta \cos| \approx 0.0805,
\]

\[
\text{Jaccard@10} \approx 0.63,
\]

and a top-1 neighbor flip rate of approximately $0.37$.

More than one third of nearest neighbors change despite the network
producing identical predictions.

\subsection{Whitening as a Canonical Gauge}

We also examined whitening as a canonical coordinate choice.

Given the covariance matrix $\Sigma$ of the representation distribution,
the whitening transformation
\[
D = \Sigma^{-1/2}
\]
maps the covariance to the identity.

In the Digits experiment the covariance eigenvalues ranged from
approximately $0.015$ to $36.6$.
After whitening all eigenvalues were approximately equal to one with
mean absolute deviation

\[
|\lambda - 1| \approx 6\times10^{-6}.
\]

Whitening removes second-order anisotropy in the representation
distribution and provides a canonical metric for representation space.

\begin{figure}[ht]
\centering
\includegraphics[width=0.6\linewidth]{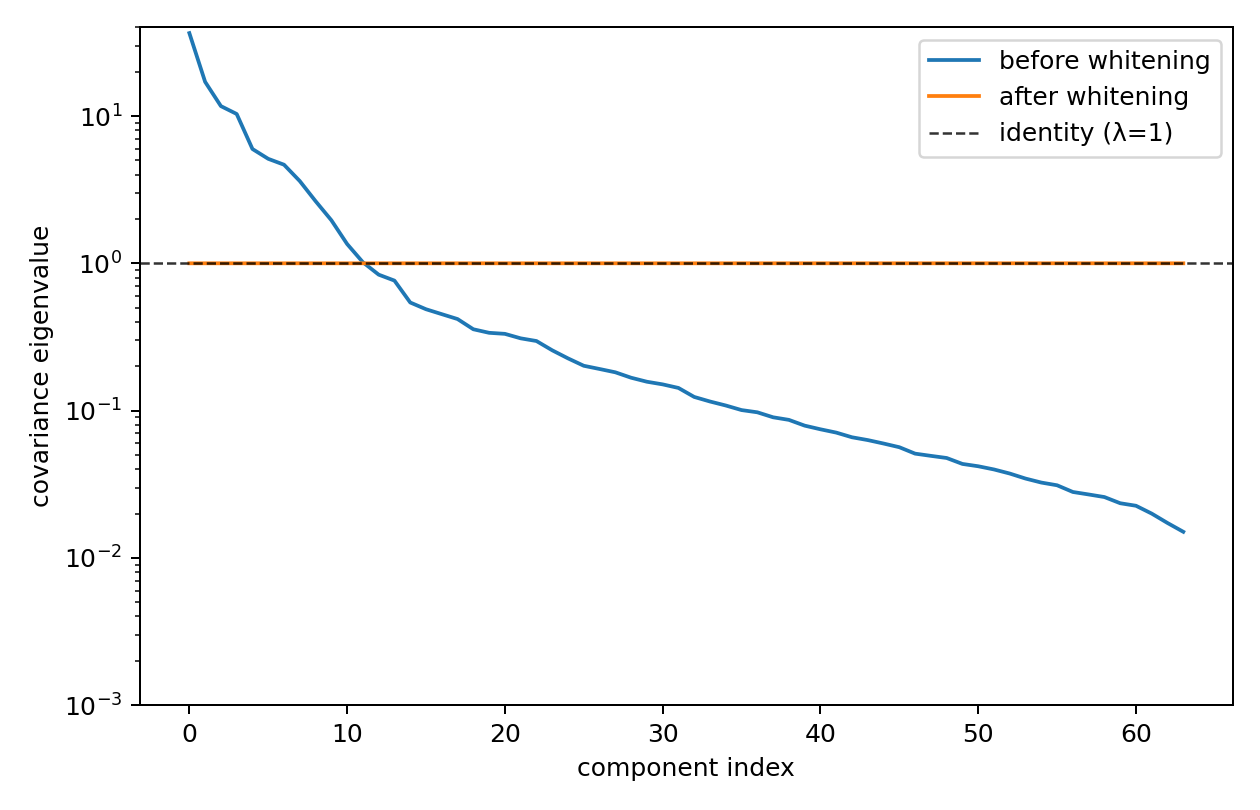}
\caption{Eigenvalue spectrum of the representation covariance matrix before
and after whitening (log scale). Whitening applies the transformation
$D=\Sigma^{-1/2}$, which fixes a canonical gauge in which the covariance
becomes the identity. After whitening the spectrum collapses to
$\lambda \approx 1$, indicating that second-order anisotropy in the
representation distribution has been removed.}
\label{fig:whitening_spectrum}
\end{figure}

The covariance eigenvalue spectrum collapses to unity after whitening
(Figure~\ref{fig:whitening_spectrum}).

\section{Discussion}

The experiments isolate a basic property of neural representations: many
geometric quantities depend on representation coordinates.
Because an invertible gauge transform can preserve the model function,
quantities such as cosine similarity or nearest-neighbor structure are
properties of a chosen gauge, not invariants of the model.

Many interpretability and embedding analyses rely on cosine similarity as a
measure of semantic or functional closeness.
Our results show that cosine-based measures can change substantially under
gauge transforms that leave predictions unchanged.

The effect grows with the condition number of the transformation.
Even moderate distortions can alter nearest-neighbor structure while
preserving model behavior, so conclusions from cosine-based analysis require
care.

One response is to use similarity measures invariant under invertible linear
transforms, including subspace-based comparison methods.
Another is to fix a canonical coordinate system; whitening is one example
because it maps the representation covariance to the identity and defines an
isotropic metric. Representation analysis should distinguish properties of
the model function
from properties of a coordinate realization. 

A practical implication is that empirical studies of representation
geometry should account explicitly for gauge dependence. Analyses based
on cosine similarity or Euclidean distance implicitly assume a fixed
coordinate system for the representation space. Because equivalent
networks may realize representations related by invertible linear
transformations, conclusions drawn from such metrics can depend on the
chosen gauge. Reporting results under a canonical coordinate choice
(such as whitening) or using comparison methods that are invariant to
linear transformations provides a more stable basis for interpreting
representation structure.

A limitation of the present study is that the experiments focus on
controlled gauge transformations inserted at a single hidden layer in
relatively small models.
Larger transformer architectures, residual pathways, and normalization
layers may introduce additional structure into the choice of practical
representation gauges.
A second limitation is that we focus primarily on second-order geometry and
cosine-based retrieval structure rather than downstream tasks that use more
complex nonlinear similarities.
These questions remain natural directions for future work.

\section{Conclusion}

Neural representation spaces can be treated as geometric objects defined
only up to invertible linear transformations.
Applying a linear transformation to a representation and compensating the
classifier leaves the network function unchanged.

Despite functional invariance, representation geometry can change
substantially.
Cosine similarity and nearest-neighbor structure may vary under gauge
transformations even when predictions are identical.

Experiments on a multilayer perceptron and a convolutional network show that
cosine distortion grows with the condition number of the transformation.
Cosine similarity is not an intrinsic property of learned
representations; it depends on coordinate choice.

Whitening provides a canonical gauge in which representation covariance
becomes isotropic.
Making gauge freedom explicit leads to cleaner interpretation of
representation geometry.

\section{Data and Code Availability}

All code, data processing scripts, trained models, and results presented in this study are publicly available for reproducibility and further research. The complete implementation is hosted as an open-source repository at:

\begin{center}
\url{https://github.com/jericho-cain/neural-representation-gauge}
\end{center}

\bibliographystyle{plainnat}
\bibliography{refs_ml}

\end{document}